\documentclass[10pt,twocolumn,letterpaper]{article}

\usepackage{wacv}              

\usepackage{comment}
\usepackage[accsupp]{axessibility}  
\usepackage{environ}         
\usepackage{calc}       
\usepackage{indentfirst}
\usepackage{multirow}
\usepackage{makecell}
\usepackage{pifont} 
\usepackage{colortbl} 
\usepackage{float}
\usepackage{subcaption}
\usepackage{algorithm}      
\usepackage{algpseudocode}  

\definecolor{wacvblue}{rgb}{0.21,0.49,0.74}
\usepackage[pagebackref,breaklinks,colorlinks,allcolors=wacvblue]{hyperref}

\def\wacvPaperID{91} 
\def\confName{WACV}
\def\confYear{2026}

\title{FlowCLAS: Enhancing Normalizing Flow-Based Anomaly Segmentation Via Contrastive Learning}

\author{
Chang Won Lee$^{1}$ \quad
Selina Leveugle$^{1}$ \quad
Paul Grouchy$^{2}$ \quad
Chris Langley$^{2}$ \quad \\
Svetlana Stolpner$^{2}$ \quad
Jonathan Kelly$^{1}$ \quad
Steven L. Waslander$^{1}$ \\
\vspace{1mm} 
$^{1}$University of Toronto \quad $^{2}$MDA Space \\
\vspace{1mm} 
{\tt\small $^{1}$\{john.lee,selina.leveugle,jonathan.kelly,steven.waslander\}@robotics.utias.utoronto.ca} \\[-0.5em]
{\tt\small $^{2}$\{paul.grouchy,chris.langley,svetlana.stolpner\}@mda.space}
}

\newcommand{\xb}{\mathbf{x}}
\newcommand{\zb}{\mathbf{z}}
\newcommand{\fb}{\mathbf{f}}
\newcommand{\nf}{f_{\theta}}
\newcommand{\bracketLR}[1]{\left( #1 \right)}
\newcommand{\union}{\cup}

\newcommand{\cmark}{\textcolor{green}{\ding{51}}}
\newcommand{\xmark}{\textcolor{red}{\ding{55}}}

\newcommand{\darkgreen}{\cellcolor{green!50}}
\newcommand{\green}{\cellcolor{green!30}}
\newcommand{\lightgreen}{\cellcolor{green!10}}

\newlength{\eqwidth}
\let\origequation=\equation
\let\origendequation=\endequation
\RenewEnviron{equation}{
  \settowidth{\eqwidth}{$\displaystyle{\BODY}$} 
  \origequation 
  \resizebox{\minof{\columnwidth}{\eqwidth}}{!}{$\displaystyle{\BODY}$} 
  \origendequation
}

\begin{document}
\maketitle
\begin{abstract}

Anomaly segmentation is an essential capability for safety-critical robotics applications that must be aware of unexpected events. Normalizing flows (NFs), a class of generative models, are a promising approach for this task due to their ability to model the inlier data distribution efficiently. However, their performance falters in dynamic scenes, where complex, multi-modal data distributions cause them to struggle with identifying out-of-distribution samples, leaving a performance gap to leading discriminative methods.
To address this limitation, we introduce FlowCLAS, a hybrid framework that enhances the traditional maximum likelihood objective of NFs with a discriminative, contrastive loss. Leveraging Outlier Exposure, this objective explicitly enforces a separation between normal and anomalous features in the latent space, retaining the probabilistic foundation of NFs while embedding the discriminative power they lack.
The strength of this approach is demonstrated by FlowCLAS establishing new state-of-the-art (SOTA) performance across multiple challenging anomaly segmentation benchmarks for robotics, including Fishyscapes Lost \& Found, Road Anomaly, SegmentMeIfYouCan-ObstacleTrack, and ALLO. Our experiments also show that this contrastive approach is more effective than other outlier-based training strategies for NFs, successfully bridging the performance gap to leading discriminative methods. 
Project page: \url{https://trailab.github.io/FlowCLAS}

\vspace{-5mm}
\end{abstract}    
\section{Introduction}
\label{sec:intro}

\begin{figure}[t]
\centering
\includegraphics[width=\columnwidth]{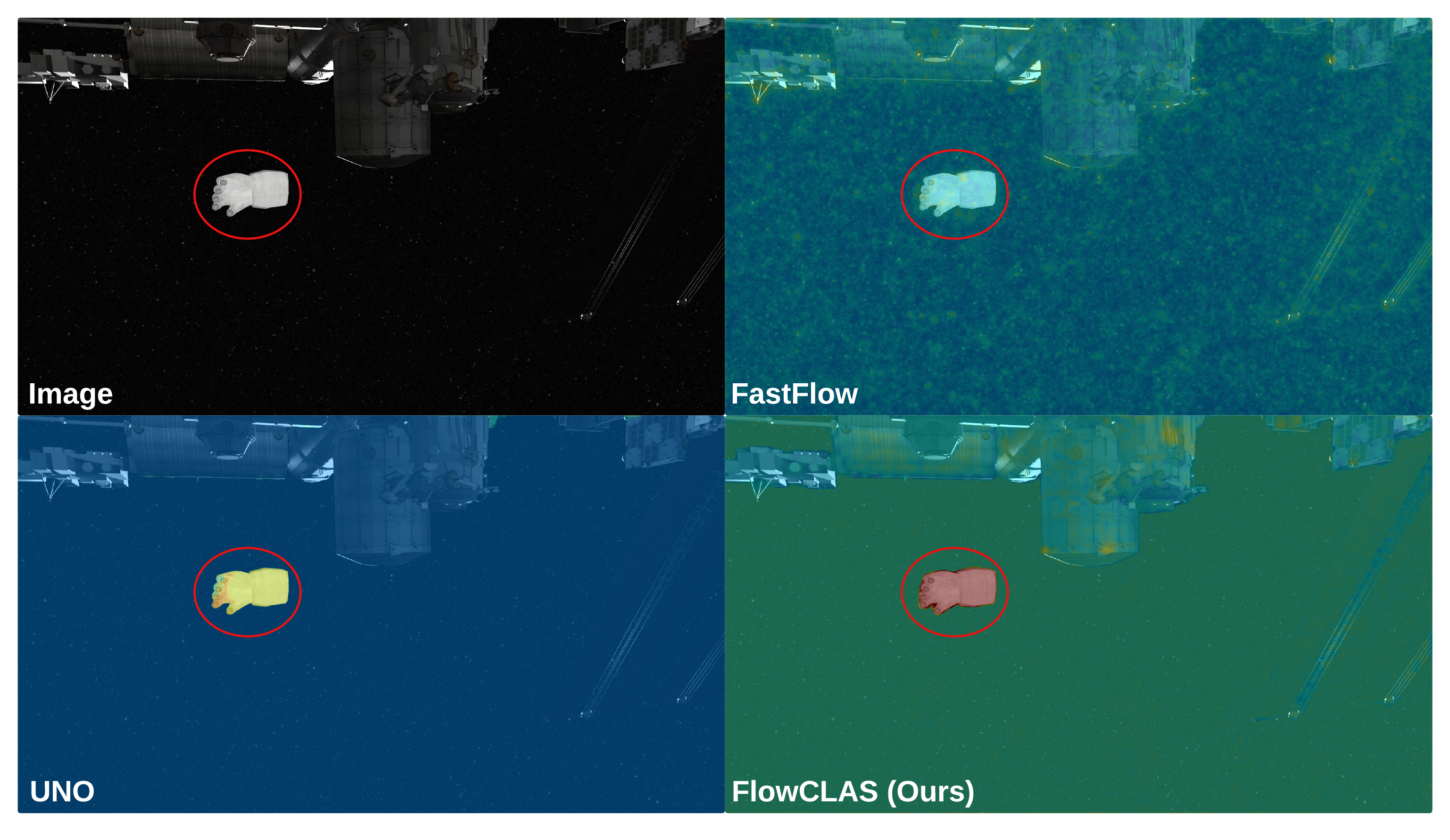}
\caption{A qualitative comparison on an ALLO \cite{leveugle_allo_2024} test image demonstrates our method's efficacy. While the unsupervised normalizing flow method, FastFlow \cite{yu_fastflow_2021}, fails to detect the anomalous glove, FlowCLAS generates a clean segmentation competitive with the supervised SOTA, UNO \cite{delic_outlier_2024}. This result validates our framework’s success in enhancing normalizing flows to bridge the performance gap with leading discriminative approaches.}
\vspace{-5mm}
\label{fig:intro_comparison}
\end{figure}

Anomaly segmentation is an essential computer vision task in robotics that focuses on detecting and localizing objects that deviate from expected patterns present in training images. By identifying out-of-distribution (OoD) visual elements that fall outside of predefined classes, anomaly segmentation complements traditional closed-set semantic segmentation and offers a more comprehensive understanding of complex visual scenes. This capability is highly valuable for safety-critical applications, of which we study two examples in this work. In autonomous driving, it can help prevent collisions with unexpected objects such as exotic animals or costumed children, while in space robotics, it can safeguard against potential damage to robotic arms from unexpected objects, thereby enhancing operational safety and efficiency \cite{leveugle_allo_2024}.


State-of-the-art (SOTA) anomaly segmentation in autonomous driving is dominated by discriminative models that learn class boundaries in a feature space, while methods for space robotics remain understudied due to the novelty of the application \cite{leveugle_allo_2024}. Although these approaches achieve strong results, they typically operate as black boxes, lacking outputs with an explicit probabilistic meaning, which is undesirable in safety-critical systems \cite{guidotti2018survey}. Furthermore, because their training objective is to separate known categories rather than to model the distribution of normal data, their robustness to outliers that differ significantly from the training samples is not guaranteed.

Normalizing flows (NFs), a class of generative models, offer a promising solution to modeling the distribution of anomaly-free data for detecting OoD samples. This approach has proven effective in controlled settings like industrial defect inspection, where images feature static, single-object views with narrow feature distributions \cite{bergmann_mvtec_2019,gudovskiy_cflow-ad_2022,yu_fastflow_2021,lei_pyramidflow_2023,tailanian_u-flow_2024}. However, the performance of NFs diminishes substantially in highly varied environments---such as autonomous driving and space robotics---where complex, multi-modal distributions of ``normal" data arise from diverse viewpoints, lighting conditions, and multiple object configurations \cite{leveugle_allo_2024}. This distributional complexity poses a significant challenge for standard NF-based methods, often leading to poor generalization and degraded performance \cite{kirichenko_why_2020}.

A key challenge for NF-based anomaly segmentation in robotics is therefore to move beyond detecting low-level anomalous patterns to identifying whole anomalous objects in dynamic scenes. This task requires a high-level semantic understanding that standard NF-based methods struggle to provide when confronted with complex, multi-modal data. To address this, we propose \textbf{FlowCLAS} (\textbf{Flow} via \textbf{C}ontrastive \textbf{L}earning for \textbf{A}nomaly \textbf{S}egmentation), a framework that leverages a frozen pre-trained backbone to extract discriminative features and trains a normalizing flow to estimate their density. Departing from prior NF-based work, FlowCLAS mixes outlier objects from an auxiliary dataset into source anomaly-free images and employs contrastive learning to separate normal and anomalous features in the latent space. This unique combination of density estimation and contrastive learning enables the reliable segmentation of anomalous objects across scenes with significant diversity, overcoming a critical limitation of existing NF-based methods in robotics.
Our key contributions are:
\begin{itemize}
    \item We introduce FlowCLAS, a novel framework that enhances normalizing flow (NF)-based density estimation with a hybrid training objective. By integrating an outlier-based contrastive loss with traditional maximum likelihood estimation, FlowCLAS learns a separable latent space to robustly segment anomalous objects in dynamic scenes.
    \item We establish through extensive ablation studies that this contrastive learning objective is critical for performance, significantly outperforming other outlier-based training strategies for NFs and proving essential for learning high-level semantic features over low-level patterns.
    \item FlowCLAS achieves state-of-the-art performance on four challenging robotics benchmarks, including Fishyscapes Lost \& Found, Road Anomaly, SegmentMeIfYouCan-ObstacleTrack, and ALLO, successfully bridging the performance gap between generative models and leading discriminative approaches.
\end{itemize}

\section{Related Work}
\label{sec:related_works}

\paragraph{Unsupervised anomaly segmentation:}

Anomaly segmentation assumes that anomalies are rare and differ significantly from normal data \cite{zimek_outlier_2017}. In the context of autonomous driving, methods are typically trained on known, in-distribution classes and are expected to detect unfamiliar objects at test time.
Early approaches treat anomaly segmentation as a post-hoc task, using uncertainty metrics from a pre-trained segmentation model, such as the entropy \cite{hendrycks_baseline_2018} or maximum softmax probability of its predictions. However, it is well documented that such closed set models can produce uncalibrated, high-confidence predictions for OoD input, limiting their reliability \cite{hendrycks_baseline_2018, gal_uncertainty_2016}. 

To address this challenge, subsequent work has focused on developing more robust anomaly indicators. Some methods refine post-hoc analysis by operating on logits instead of softmax probabilities, such as by using standardized max logits \cite{jung_standardized_2021} or training an additional network on top of the logits to model uncertainty \cite{gudovskiy_concurrent_2023}. Other approaches retrain the segmentation network with specialized objectives to better distinguish the known distribution, using techniques such as energy-based models \cite{tian_pixel-wise_2022} or hybrid classifiers \cite{avidan_densehybrid_2022}. For example, SynBoost \cite{di_biase_pixel-wise_2021} combines softmax-based uncertainty with reconstruction error from an image re-synthesis network, leveraging both semantic and geometric discrepancies. Despite these advances, the performance of purely unsupervised methods is inherently limited by the absence of outlier knowledge during training.

\vspace{-3mm}

\paragraph{Normalizing flows for density estimation:}

Normalizing flows (NFs) are a well-established tool for anomaly segmentation, particularly in controlled domains such as industrial inspection and medical imaging. The standard approach, pioneered by methods like FastFlow \cite{yu_fastflow_2021}, involves training an NF to model the probability distribution of features from a backbone network and assigning low likelihood scores to anomalous regions. Building on this, PyramidFlow \cite{lei_pyramidflow_2023} introduced a hierarchical architecture in which fine-grained feature flows are conditioned on outputs from coarser scales, integrating global semantic context with local details. AE-FLOW \cite{zhao_ae-flow_2022} integrates an NF into an autoencoder's bottleneck, yielding a robust anomaly score from two complementary signals: latent likelihood and reconstruction error. Despite these architectural variations, these methods share a common strategy---that is, modeling the distribution of inlier data.

However, applying this density estimation paradigm to autonomous driving and space robotics remains a significant challenge due to scene complexity and high data variance. Direct likelihood maximization with NFs often causes the model to focus on low-level pixel statistics rather than the high-level semantic content required to identify anomalous objects \cite{kirichenko_why_2020}. Consequently, the most relevant works in this domain re-purpose NFs for auxiliary tasks rather than for direct feature density modeling. For instance, FlowEneDet \cite{gudovskiy_concurrent_2023} applies a flow to energy scores derived from classifier logits to better model uncertainty, while other methods use NFs to synthesize outlier data for supervised training \cite{avidan_densehybrid_2022}. This leaves the challenge of building an effective, direct NF-based density estimator for complex scenes largely unaddressed.

\vspace{-4mm}

\paragraph{Supervised anomaly segmentation with outliers:}
A prominent line of research enhances anomaly segmentation by leveraging Outlier Exposure (OE) \cite{hendrycks_deep_2018}, which integrates OoD data into the training process. Early work in this area for road anomaly segmentation, such as Meta-OoD \cite{chan_entropy_2021}, mixed full OoD images from datasets like COCO \cite{lin_microsoft_2014} into the training set, using an entropy maximization objective for outlier regions. Subsequent methods created more contextually coherent anomalies by synthetically inserting outlier objects into normal scenes.
Building on this foundation, recent approaches have developed sophisticated, discriminative training objectives. PEBAL \cite{tian_pixel-wise_2022} combines OE with energy-based learning and abstention learning, while DenseHybrid \cite{avidan_densehybrid_2022} trains a hybrid segmentation network to explicitly separate known classes from an explicit outlier class through likelihood training. UNO \cite{delic_outlier_2024} also treats outliers as an additional class, fusing inlier uncertainty with a novel negative objectness score. Notably, RPL \cite{liu_residual_2023} introduces a contrastive learning objective to enforce a clear separation between inlier and outlier feature representations, demonstrating a significant boost in discriminative power.

In this work, we introduce a hybrid framework that enhances NF-based density estimation with a discriminative, contrastive objective. Our approach is designed to retain the generative model's core strength---its ability to model the distribution of normal data for robust anomaly segmentation---while integrating the power of contrastive learning to actively structure the latent space. By explicitly enforcing a separation between inlier and outlier representations, we address a critical failure mode of NFs, which can assign high likelihood to anomalous samples \cite{kirichenko_why_2020}. The result is a system that unites the probabilistic interpretability of a generative model with the discriminative power required for anomaly segmentation in complex, dynamic scenes.

\section{Methodology}
\label{sec:methodology}

Our methodology builds upon the use of normalizing flows for feature density estimation. We first review this foundational approach and its application to anomaly segmentation via maximum likelihood estimation. We then introduce a novel training objective that enhances this framework by integrating contrastive learning with auxiliary outlier samples. This hybrid objective is designed to improve the model's discriminative power, enabling a clearer separation between normal and anomalous features.

\subsection{Normalizing flow overview}

A normalizing flow is a type of generative neural network that learns a differentiable, invertible mapping with parameters $\theta$ ${\mathbf{f}_{\theta}: X \rightarrow Z}$ between the input space $X$ and a latent space $Z$ \cite{papamakarios_normalizing_2021}. This diffeomorphism transforms input samples ${\mathbf{x} \in X}$ (\eg image pixels or features) into latent samples ${\mathbf{z} \in Z}$, whose distribution is typically modeled by a Multivariate Gaussian distribution (MGD). The invertibility of this formulation enables exact likelihood computation via the change of variables formula, yielding $p_X(\mathbf{x})$ \cite{papamakarios_normalizing_2021}:
\begin{equation}
p_X(\xb)=p_Z(\zb)\left|\operatorname{det}\left(\frac{\partial \zb}{\partial \xb}\right)\right|,
\label{eq:change_of_variable}
\end{equation}
where ${\frac{\partial \zb}{\partial \xb}}$ is the Jacobian of the invertible mapping $\fb_{\theta}$ and ${\zb \sim \mathcal{N}(\mathbf{\mu}, \mathbf{\Sigma})}$. Here, $\mathbf{\mu}$ and $\mathbf{\Sigma}$ are  parameters of the MGD which can be fixed (typically zero mean and unit covariance) or also learned.
The network is then trained to maximize the log-likelihood of the input as follows:
\begin{equation}
\log p_X(\xb) = \log p_Z\left( \zb \right)+\log \left|\operatorname{det}\left(\frac{\partial \zb}{\partial \xb}\right)\right|
\end{equation}

\subsection{Pseudo-anomalies with Outlier Exposure}

We describe a method to generate anomalous images by pasting random objects from an auxiliary dataset into anomaly-free training images, similar to \cite{di_biase_pixel-wise_2021,avidan_densehybrid_2022,liu_residual_2023}. Let ${\mathcal{D}^{in}=\left\{\mathbf{I}_i^{in}\right\}_i^{\left|\mathcal{D}^{in}\right|}}$ denote the anomaly-free training dataset, where $\mathbf{I}_i^{in}$ is an input image. We also define an auxiliary dataset ${\mathcal{D}^{out}=\left\{\left(\mathbf{I}_i^{out}, \mathbf{M}_i^{out}\right)\right\}_i^{\left|\mathcal{D}^{out}\right|}}$, where $\mathbf{M}_i^{out}$ is a binary mask separating foreground objects from the background.
To create anomalous images, we copy-paste objects from $\mathcal{D}^{out}$ into $\mathcal{D}^{in}$ using the binary masks, resulting in a mixed dataset ${\mathcal{D}^{mix}=\left\{\left(\mathbf{I}_i^{mix}, \mathbf{M}_i^{mix}\right)\right\}_i^{\left|\mathcal{D}^{mix}\right|}}$. The mixed image $\mathbf{I}_i^{mix}$ is computed as:
\begin{equation}
\mathbf{I}_i^{mix}=(\mathbf{1} - \mathbf{M}_i^{mix}) \odot \mathbf{I}_i^{in}+\mathbf{M}_i^{mix} \odot \mathbf{I}_i^{out}
\end{equation}
where $\mathbf{M}_i^{mix}$ is the resulting binary mask after randomly scaling the anomalous object and pasting it into a random location of the original image.

\subsection{Normalizing flow for anomaly segmentation}
\begin{figure*}[t]
  \centering
  \includegraphics[width=0.9\textwidth]{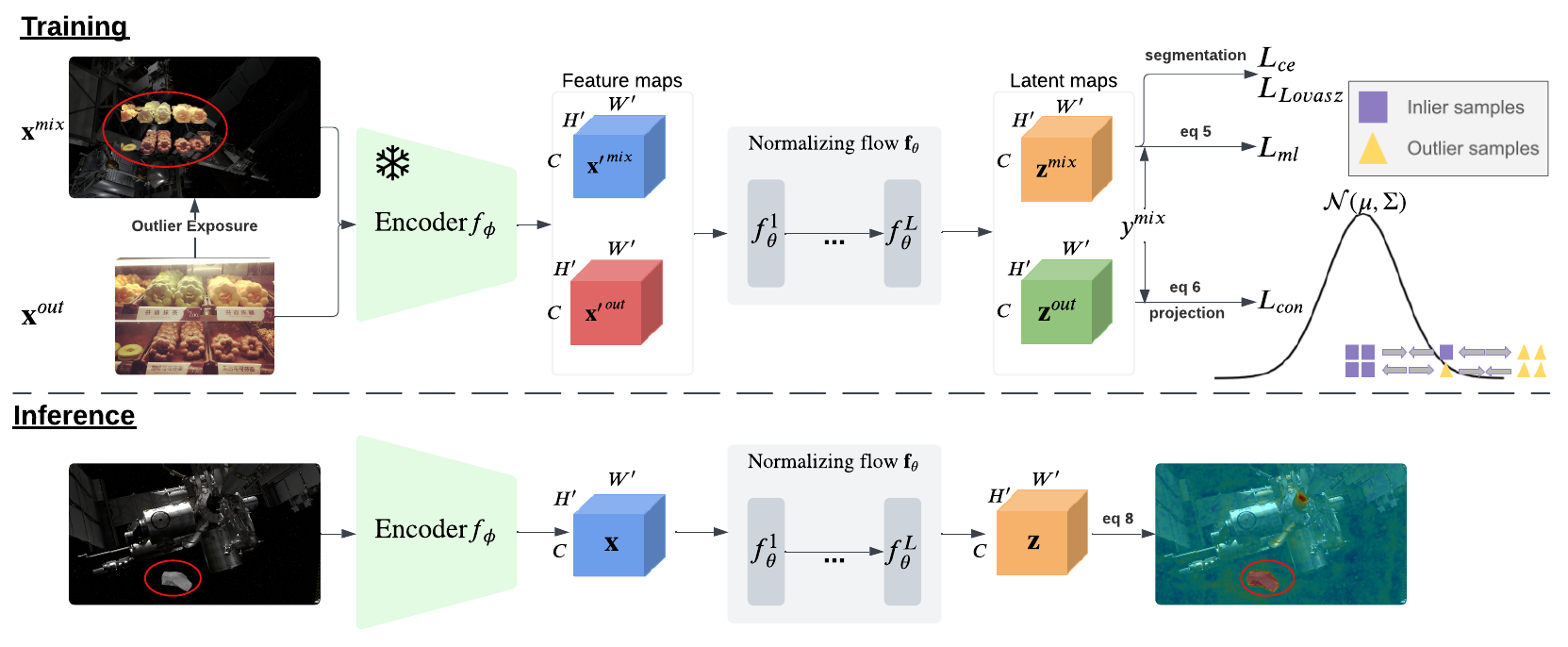}
   \caption{
   The FlowCLAS framework operates in two stages. During training, a frozen vision encoder $\mathbf{f}_{\phi}$ extracts features from a mixed-content image $\xb^{mix}$ and an auxiliary outlier image $\xb^{out}$. A normalizing flow network $\mathbf{f}_{\theta}$ then maps these features to a latent space, producing $\zb^{\{mix,out\}}$. The model is optimized via a hybrid objective: (1) a maximum likelihood loss ($\mathcal{L}_{ml}$) pushes the latent samples $\zb$ corresponding to normal regions toward a base Multivariate Gaussian distribution, and (2) a contrastive loss ($\mathcal{L}_{con}$) enforces a separation between the latent representations of normal and anomalous features in the projection space. During inference, the normalizing flow computes a likelihood-based anomaly score map for a given test image, localizing regions that deviate from the learned distribution of normal data.
   }
\vspace{-3mm}
\label{fig:methodology:flowclas_overview}
\end{figure*}

Let us denote a frozen pre-trained feature extraction network such as a vision foundation model as $f_{\phi}$ that extracts 2D feature maps ${\xb \in \mathbb{R}^{H \times W \times C}}$ from RGB images. In the case of a vision transformer, the tokens are reshaped to a 2D grid to ensure spatial relationships are preserved. Then, these maps are fed into a differentiable 2D flow $\mathbf{f}_{\theta}$ that generates latent maps of the same dimensions $\zb \in \mathbb{R}^{H \times W \times C}$. 

As seen in \cref{fig:methodology:flowclas_overview}, this image-to-latent transformation process is repeated for a mixed image and an outlier image to produce latent maps $\zb^{mix}$ and $\zb^{out}$. The outlier image is provided in addition to the mixed image to produce more samples for the contrastive set described in \cref{sec:methodology:project_lower_space}.

In practice, the mapping ${\mathbf{f}_{\theta}(\cdot)}$ is implemented as a sequence of $L$ invertible blocks as $\mathbf{f}_{\theta} = \nf^1 \circ \nf^2 \circ \cdots \circ \nf^L$. Each block is a series of invertible operations:
\begin{equation}
    \zb^l = \operatorname{Affine}\bracketLR{\operatorname{RandPermute}\bracketLR{\operatorname{Actnorm}\bracketLR{\zb^{l-1}}}},
\end{equation}
where $\operatorname{Actnorm}$ is the activation normalization from \cite{kingma_glow_2018}, $\operatorname{RandPermute}$ is a random channel-wise permutation and $\operatorname{Affine}$ is an affine coupling layer \cite{dinh_density_2017} with a residual subnetwork.

The normalizing flow $\mathbf{f}_{\theta}$ is trained to maximize the log-likelihoods, or alternatively minimize the negative log-likelihoods, of \emph{non-anomalous} features via \cref{eq:change_of_variable}:
\begin{equation}
\begin{aligned}
L_{ml} &= \bracketLR{{\mathbf{y}}^{mix} - \mathbf{1}} \log p_{X}\bracketLR{{\xb}^{mix}} \\
&= \bracketLR{{\mathbf{y}}^{mix} - \mathbf{1}} \left[ \log p_Z\left( \zb^{mix} \right)+\log \left|\operatorname{det}\left(\frac{\partial \zb^{mix}}{\partial {\xb}^{mix}}\right)\right| \right],
\end{aligned}
\label{eq:method:nll}
\end{equation}
where ${\mathbf{y}}^{mix} \in \mathbb{R}^{H \times W}$ is the ground-truth binary mask interpolated to match the latent map's dimensions. In this formulation, the latent distribution $p_Z$ is a multivariate Gaussian $\mathcal{N}(\mathbf{\mu}, \mathbf{\Sigma})$ with a learnable mean $\mathbf{\mu} \in \mathbb{R}^C$ and a diagonal covariance matrix $\mathbf{\Sigma} \in \mathbb{R}^{C \times C}$. These parameters are optimized jointly with the flow.

\subsection{Pseudo-outliers for enhanced latent separation}

In the traditional maximum likelihood framework, normalizing flow networks are trained to assign high likelihood scores to inliers, with the implicit assumption that outlier features will naturally be projected far from inlier latent samples, resulting in low likelihood scores. However, this assumption can fail in practice. To address this, we introduce a contrastive learning objective using Outlier Exposure \cite{hendrycks_deep_2018}. This approach explicitly encourages the model to map outlier features to low-likelihood regions of the latent space, distant from the high-likelihood regions occupied by inliers, thereby enhancing the network's ability to distinguish between the two classes. 
is it common to have tables at the bottom of the page rather than the top of the page in computer vision conference papers?

\subsubsection{Projection to lower-dimensional space}
\label{sec:methodology:project_lower_space}

Our model first encodes images into latent feature maps. 
Let $\mathcal{Z}^{mix}$ be the set of latent maps $\zb^{mix}$ from a batch of mixed-content images $\mathcal{B}^{mix}$. It is defined as $\mathcal{Z}^{mix} = \left\{ \zb^{mix}=f_\theta \circ f_\phi\left(\mathbf{I}^{mix}\right), \forall \mathbf{I}^{mix} \in \mathcal{B}^{mix} \right\}$. Similarly, define $\mathcal{Z}^{out}$ for a batch of pure outlier images $\mathcal{B}^{out}$. 
Following standard practice \cite{he_momentum_2020,chen_simple_2020,khosla_supervised_2020} in contrastive learning, we project these latent maps to a lower dimension through a projection head comprising a $1 \times 1$ convolutional layer and channel-wise L2 normalization, to obtain the project sets $\tilde{\mathcal{Z}}^{mix}$ and $\tilde{\mathcal{Z}}^{out}$. This projection maps latent representations into a lower-dimensional space to enhance their discriminative power and reduce redundancy for contrastive learning.

\subsubsection{Supervised contrastive learning with outliers}

Following \cite{liu_residual_2023}, we first build an anchor set $\mathcal{A}$, by using the binary masks of $\mathcal{B}^{mix}$ to randomly sample an equal number of inlier and outlier latent vectors from the set of projected latent maps $\tilde{\mathcal{Z}}^{mix}$.
We then form an auxiliary outlier set $\mathcal{B}$ by repeating the sampling process with $\tilde{\mathcal{Z}}^{out}$ using the binary masks of $\mathcal{B}^{out}$. Finally, we form a contrastive set $\mathcal{C}$ by combining the two sets as: $\mathcal{C} = \mathcal{A} \union \mathcal{B}$.

For a given anchor vector $\zb_i \in \mathcal{A}$, the set of positive samples $\mathcal{C}^{+}(\zb_i)$ contains all other vectors in $\mathcal{C}$ from the same class (inlier or outlier). The set of negative samples $\mathcal{C}^{-}(\zb_i)$ comprises all vectors from the other class. We then apply the InfoNCE contrastive loss \cite{oord2018representation} which aims to maximize intra-class similarity and inter-class dissimilarity:
\begin{equation}
\begin{aligned}
    &L_{con} = \\
    &\sum_{\zb_i \in \mathcal{A}} \sum_{\zb_i^{+} \in \mathcal{C}^{+}\left(\zb_i\right)} -\log \frac{\exp \left(\zb_i \cdot \zb_i^{+} / \tau\right)}{\exp \left(\zb_i \cdot \zb_i^{+} / \tau\right)+\sum_{\zb_i^{-} \in \mathcal{C}^{-}\left(\zb_i\right)} \exp \left(\zb_i \cdot \zb_i^{-} / \tau\right)},
\end{aligned}
\end{equation}
where $\tau$ denotes the temperature hyperparameter. This loss function encourages the flow to learn a latent space where embeddings from the same class are drawn closer, while those from different classes become more separated.

\subsection{FlowCLAS training and inference}

To further enhance the discriminative power of the latent space, we introduce an auxiliary segmentation objective. We attach a lightweight segmentation head to the latent features $\zb^{mix}$ and train it using a combination of cross-entropy ($L_{ce}$) and Lovasz-Softmax ($L_{Lovasz}$) losses.

The total loss function used to train the normalizing flow network is defined as:
\begin{equation}
    L_{NF} = \alpha L_{ml} + L_{con} + L_{ce} + L_{Lovasz},
\end{equation}
where is $\alpha$ is a tunable hyperparameter. 

At inference, the projection head is omitted. For a given query image, we first obtain its 2D latent map $\zb \in \mathbb{R}^{H \times W \times C}$. 
To produce a 2D anomaly score map, we compute an anomaly score for each latent vector $\zb_{i,j} \in \mathbb{R}^{C}$ at spatial coordinates $(i,j)$, where $i \in [0, H)$ and $j \in [0, W)$:
\begin{equation}
\operatorname{score}(\zb_{i,j}) = 1 - \exp{\left (\frac{1}{C} \log p_Z(\zb_{i,j}) \right)}
\end{equation}

This formulation posits that outliers yield lower log-likelihood scores compared to inliers. The final anomaly score map is interpolated to match the resolution of the input image.

\subsection{Mask-based Anomaly Score Smoothing}

Pixel-wise anomaly scoring can produce imprecise object boundaries, particularly when the score map is upscaled from a low-resolution latent space. Furthermore, local artifacts---originating from the feature extractor (\eg positional encoding noise in Vision Transformers ) or noise from the input image---can cause score inconsistencies within a single object instance. 

To enforce instance-level consistency in the anomaly scores, we use an external class-agnostic mask predictor to segment the input image into distinct regions. For each mask, we construct a histogram with $N$ bins from the anomaly scores. We then identify the bin with the highest frequency and compute its mean score. Finally, we update all pixels within the mask whose scores do not fall into this dominant bin with the computed mean.

\section{Experiments}
\label{sec:experiments}

\subsection{Experimental setup}
\paragraph{Datasets:}
We conduct our road anomaly segmentation experiments on the Lost\&Found subset of Fishyscapes (FS-L\&F) \cite{blum_fishyscapes_2019}, the Road Anomaly dataset \cite{lis_detecting_2019}, and the SegmentMeIfYouCan (SMIYC) benchmark \cite{chan_segmentmeifyoucan_2021}. FS-L\&F offers 100 urban driving images with anomalies for validation from the Lost and Found dataset \cite{pinggera_lost_2016}. Similarly, Road Anomaly provides 60 validation anomalous images of various road scenes. The SMIYC benchmark has two subsets, AnomalyTrack and ObstacleTrack which contain anomalous objects of various sizes in different contexts. In these experiments, we use the Cityscapes dataset \cite{cordts_cityscapes_2016} for training, which contains 2,975 images across 19 object classes. 
We also evaluate our method's performance on an extended version of the ALLO dataset \cite{leveugle_allo_2024} for space anomaly segmentation. Using the open-source code, we generated 29,200 anomaly-free training images and 22,324 test images, each accompanied by a 9-class mask indicating the background, various parts of the ISS and the anomalous object, if present.
For outlier exposure, we copy-paste objects from the COCO dataset \cite{lin_microsoft_2014}, following \cite{liu_residual_2023,chan_entropy_2021,di_biase_pixel-wise_2021}. 

\vspace{-3mm}

\paragraph{Evaluation metrics:}
We evaluated pixel-level performance using two binary classification metrics: 
Area Under Precision Recall Curve (AUPRC), 
and the False Positive Rate (FPR) at a True Positive Rate (TPR) of 95\% ($\text{FPR}_{95}$). 
The AUPRC metric is prioritized as it effectively measures anomaly detection capability in scenarios with significant class imbalance, such as large images with small anomalous objects.
The $\text{FPR}_{95}$ metric is also critical due to its significance in robotics applications, where achieving a high TPR is crucial while minimizing the number of false positives.

\vspace{-3mm}

\paragraph{Implementation details:}

For feature extraction in our main experiments, we use DINOv2-L \cite{oquab_dinov2_2023}, which has been fine-tuned for semantic segmentation using the Rein framework \cite{wei_stronger_2024} (denoted DINOv2-L-Rein). Class-agnostic mask predictions are generated using SAM 2 \cite{ravi_sam_2024}. Our normalizing flow module is constructed with $L=16$ coupling blocks and we use a projection head that reduces feature dimensionality to $C'=128$. For score smoothing, we employ a histogram with $N=20$ bins. In our ablation studies, we replace the feature extractor with DINOv2-B-Rein (base variant) and deactivate score smoothing, unless otherwise noted.
FlowCLAS was trained using a cosine learning rate scheduler with linear warmup and a batch size of 16. We set the maximum learning rate to $1 \times 10^{-6}$. The temperature scale was set to $\tau=0.13$ and the weight for the maximum likelihood loss, $\alpha$, was set to $1 \times 10^{-6}$ to ensure its magnitude was comparable to the other loss components. We trained for 50 epochs on the ALLO dataset and 600 epochs on Cityscapes.
Our implementation is built on the Anomalib library \cite{akcay_anomalib_2022} used by the ALLO benchmark \cite{leveugle_allo_2024}. The normalizing flow components are implemented using the FrEIA \cite{freia}. For supervised anomaly segmentation baselines on the ALLO dataset, we used the original source code provided by the respective authors. Further details are available in the supplementary material.

\subsection{Experimental results}

\subsubsection{Road anomaly segmentation results}

\begin{table}[t]
    \centering
    \resizebox{\columnwidth}{!}{%
    \begin{tabular}{@{}c|cc|cc@{}}
    \toprule
    \multirow{2}{*}{Method} & \multicolumn{2}{c|}{FS-L\&F} & \multicolumn{2}{c}{Road Anomaly} \\
    \cline{2-5}
    & AUPRC $\uparrow$ & $\text{FPR}_{95}$ $\downarrow$ & AUPRC $\uparrow$ & $\text{FPR}_{95}$ $\downarrow$ \\
    \midrule
    FastFlow + DINOv2-L-Rein \cite{yu_fastflow_2021} & 
    8.9 & 94.4 & 57.1 & 91.8 \\
    SynBoost \cite{di_biase_pixel-wise_2021} & 
    60.6 & 31.0 & 41.8 & 59.7 \\
    DenseHybrid \cite{avidan_densehybrid_2022} &  
    69.8 & 5.1 & - & - \\
    PEBAL \cite{tian_pixel-wise_2022} &  
    58.8 & 4.8 & 45.1 & 44.6 \\
    RbA \cite{nayal_rba_2023} &  
    70.8 & 6.3 & 85.4 & \lightgreen6.9 \\
    RPL \cite{liu_residual_2023} &  
    70.6 & 2.5 & 71.6 & 17.7 \\
    EAM \cite{grcic_advantages_2023} & 
    81.5 & 4.2 & 69.4 & 7.7 \\
    RWPM \cite{zeng_random_2025} &  
    71.2 & 6.1 & \lightgreen87.3 & \green5.3 \\
    UNO \cite{delic_outlier_2024} &  
    \lightgreen81.8 & \green1.3 & \green88.5 & 7.4 \\
    UNO + DINOv2-L-Rein \cite{delic_outlier_2024,wei_stronger_2024} & \green85.6 & \lightgreen1.5 & 75.3 & 10.9 \\
    FlowCLAS (ours) & 
    \darkgreen88.8 (+3.2) & \darkgreen0.7 (-0.6) & \darkgreen93.0 (+4.5) & \darkgreen3.3 (-2.0) \\
    \bottomrule
    \end{tabular}
    }
    \caption{Comparison with baselines on Fishyscapes Lost\&Found (FS-L\&F) and Road Anomaly validation sets. The top three results are highlighted in dark green, green and light green in order.}
    \label{table:road_val_results}
\end{table}

\begin{table}[t]
    \centering
    \resizebox{\columnwidth}{!}{%
    \begin{tabular}{@{}c|cc|cc@{}}
    \toprule
    \multirow{2}{*}{Method} & \multicolumn{2}{c|}{Anomaly Track} & \multicolumn{2}{c}{Obstacle Track} \\
    \cline{2-5}
    & AUPRC $\uparrow$ & $\text{FPR}_{95}$ $\downarrow$ & AUPRC $\uparrow$ & $\text{FPR}_{95}$ $\downarrow$ \\
    \midrule
    PEBAL \cite{tian_pixel-wise_2022} & 49.1 & 40.8 & 5.0 & 12.7 \\
    SynBoost \cite{di_biase_pixel-wise_2021} & 56.4 & 61.9 & 71.3 & 3.2 \\
    DenseHybrid \cite{avidan_densehybrid_2022} & 78.0 & 9.8 & 87.1 & \green0.2 \\
    RPL \cite{liu_residual_2023} & 83.5 & 11.7 & 85.9 & 0.6 \\
    RbA \cite{nayal_rba_2023} & 90.9 & 11.6 & 91.8 & 0.5 \\
    RWPM \cite{zeng_random_2025} & 92.0 & 10.2 & \green93.3 & \lightgreen0.3 \\
    EAM \cite{grcic_advantages_2023} & \lightgreen93.8 & \green4.1 & 92.9 & 0.5 \\
    UNO \cite{delic_outlier_2024} & \darkgreen96.3 & \darkgreen2.0 & \lightgreen93.2 & \green0.2 \\
    FlowCLAS (ours) & \green94.3 (-2.0) & \lightgreen6.6 (+4.6) & \darkgreen94.2 (+0.9) & \darkgreen0.1 (-0.1) \\
    \bottomrule
    \end{tabular}
    }
    \caption{Results on the SegmentMeIfYouCan benchmark \cite{chan_segmentmeifyoucan_2021}.}
    \label{table:smiyc_results}
    \vspace{-5mm}
\end{table}

FlowCLAS establishes a new state-of-the-art on multiple anomaly segmentation benchmarks. As detailed in \cref{table:road_val_results}, FlowCLAS consistently outperforms recent methods across all metrics on two major road anomaly benchmarks. In addition, it surpasses FastFlow \cite{yu_fastflow_2021} even when both models use the same powerful DINOv2-L-Rein feature extractor, marking a significant advancement for normalizing flow-based density estimation in this domain. The method's robust performance extends to the SMIYC benchmark (\cref{table:smiyc_results}), where it achieves state-of-the-art results on the ObstacleTrack subset and remains highly competitive on the AnomalyTrack subset. These findings highlight the effectiveness of our discriminative contrastive learning framework in enhancing normalizing flows for complex anomaly segmentation tasks.

\subsubsection{Space anomaly segmentation results}

\begin{table}[t]
    \centering
    \resizebox{\columnwidth}{!}{%
    \begin{tabular}{@{}c|cc@{}}
        \toprule
        Method & AUPRC $\uparrow$ & $\text{FPR}_{95}$ $\downarrow$ \\
        \midrule
        Rev. Dist. \cite{deng_anomaly_2022} & 8.7 & 97.9 \\
        STFPM \cite{wang_student-teacher_2021} & 14.1 & 47.6 \\
        FastFlow \cite{yu_fastflow_2021} & 29.2 & 56.5 \\
        FastFlow+DINOv2-L-Rein \cite{yu_fastflow_2021} & 13.5 & 65.2 \\
        RbA \cite{nayal_rba_2023} & 50.5 & 75.5 \\
        RWPM \cite{zeng_random_2025} & \lightgreen51.6 & \lightgreen52.5 \\
        UNO \cite{delic_outlier_2024} & \green80.8 & \darkgreen5.4 \\
        FlowCLAS (ours) & \darkgreen88.4 (+7.6) & \green6.6 (+1.2) \\
        \bottomrule
    \end{tabular}
    }
    \caption{Comparison with baselines on the ALLO test set. All methods were evaluated using 5,000 threshold bins.}
    \label{table:allo_results}
    \vspace{-5mm}
\end{table}

We further assess our method on anomaly segmentation for space robotics using the ALLO benchmark \cite{leveugle_allo_2024}. We benchmark FlowCLAS against two types of methods: (i) unsupervised approaches from industrial and medical domains that do not leverage outlier supervision \cite{yu_fastflow_2021,wang_student-teacher_2021,deng_anomaly_2022}; and (ii) outlier-supervised Mask2Former-based methods such as RbA \cite{nayal_rba_2023}, RWPM \cite{zeng_random_2025}, and UNO \cite{delic_outlier_2024}, the previous SOTA on FS-L\&F and Road Anomaly.

Consistent with the ALLO benchmark study \cite{leveugle_allo_2024}, methods designed for narrow, fixed-perspective domains like industrial inspection generally underperform on ALLO's dynamic scenes (\cref{table:allo_results}, rows 1-4). More notably, we observe that the FastFlow baseline performs worse when equipped with a stronger DINOv2-L-Rein backbone compared to a standard one. This counter-intuitive result suggests a mismatch between the highly expressive transformer-based features and the normalizing flow model. A potential cause includes the DINOv2 features creating an overly complex, high-dimensional inlier distribution that is difficult for the flow to model. With the maximum likelihood objective alone, the flow model may overfit to fine-grained details in the rich features, failing to capture the broader semantic context necessary for robust anomaly segmentation.

\begin{figure*}[ht]
\centering
\includegraphics[width=0.9\linewidth]{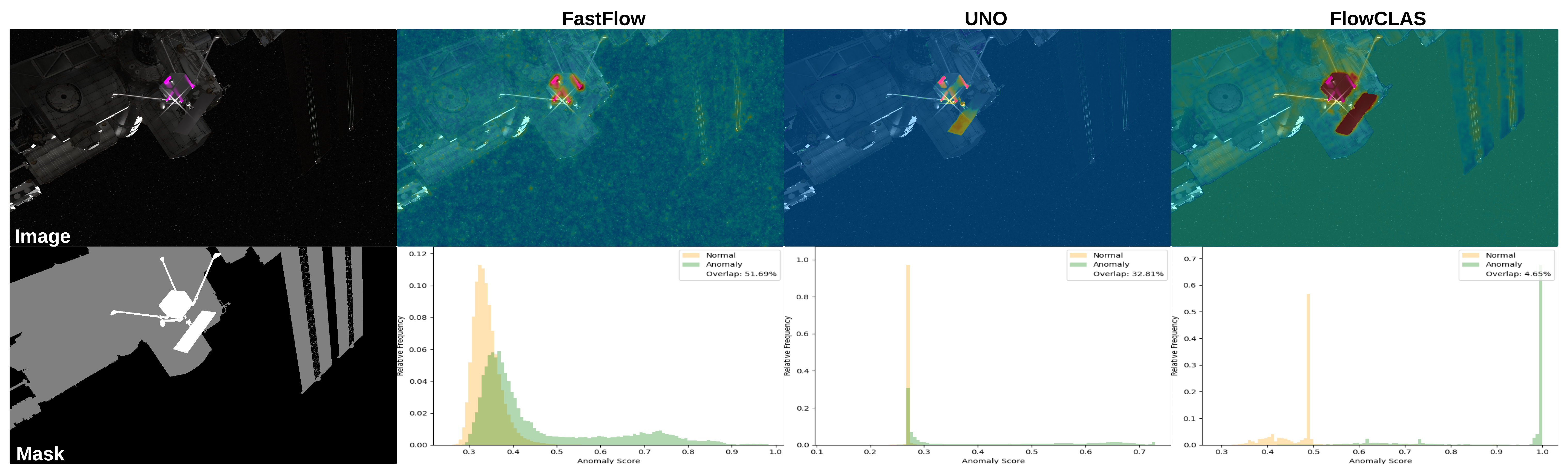}
\caption{Predicted heatmaps (top) and anomaly score histograms (bottom) for an ALLO test image. While the leading unsupervised method, FastFlow \cite{yu_fastflow_2021}, fails to detect the anomalous helicopter, and the supervised SOTA, UNO \cite{delic_outlier_2024}, cleanly detects the main body, FlowCLAS provides a more complete segmentation that captures the entire object structure. The histograms below visually illustrate this, showing the superior anomaly score separation between the two classes achieved by FlowCLAS compared to both methods.}
\label{fig:results:allo}
\vspace{-3mm}
\end{figure*}

In contrast, FlowCLAS demonstrates a significant performance leap. It substantially outperforms both variants of FastFlow, the leading unsupervised normalizing flow method, a success clearly illustrated by the qualitative results in \cref{fig:results:allo}. In this challenging low-light scenario, FlowCLAS accurately segments the entire anomalous helicopter, including its fine-grained structures. Furthermore, the failure of FastFlow---which only detects the visually distinct pink region of the object---highlights a key weakness of traditional NF approaches: a tendency to focus on low-level patterns rather than high-level object semantics. This is precisely the limitation our hybrid training objective is designed to overcome. FlowCLAS's superior discriminative power is further evidenced by the dramatically reduced overlap between the anomaly score histograms of inlier and outlier samples, indicating less confusion between the two classes. Ultimately, by achieving state-of-the-art AUPRC and the second-best FPR$_{95}$, our method reduces the performance gap to fully supervised approaches, underscoring the efficacy of our contrastive learning framework in enhancing normalizing flows for anomaly segmentation.

\subsection{Ablation studies}
\label{sec:ablations}

\subsubsection{Effect of outlier exposure on normalizing flow}
{
\setlength{\textfloatsep}{5pt}
\begin{table}[t]
    \centering
    \resizebox{\columnwidth}{!}{%
    \begin{tabular}{@{}c|c|c|c|cc|cc@{}}
    \toprule
    \multirow{2}{*}{Method Type} & \multirow{2}{*}{OE} & \multirow{2}{*}{\# cls} & \multirow{2}{*}{$L_{con}$} & \multicolumn{2}{c|}{FS-L\&F} & \multicolumn{2}{c}{Road Anomaly} \\
    \cline{5-8} & & & & AUPRC $\uparrow$ & $\text{FPR}_{95}$ $\downarrow$ & AUPRC $\uparrow$ & $\text{FPR}_{95}$ $\downarrow$ \\
    \midrule
    Baseline & \xmark & \xmark & \xmark & 62.6 & 7.2 & 57.8 & 23.7 \\
    Segmentation & \xmark & 19 & \xmark & 13.0 & 66.3 & 48.1 & 59.8 \\
    \midrule
    Binary seg. & \cmark & 2 & \xmark & 75.4 & 2.4 & 83.9 & 8.2 \\
    K+1 seg. & \cmark & 20 & \xmark & 75.8 & 3.9 & 86.0 & 7.0 \\
    Outlier min. & \cmark & \xmark & \xmark & 77.9 & 2.9 & 87.5 & 9.9 \\
    \midrule
    CL only & \cmark & \xmark & \cmark & 79.6 & 1.6 & 88.4 & 5.5 \\
    K seg. + CL & \cmark & 19 & \cmark & 80.7 & 1.4 & 90.6 & 5.0 \\
    \bottomrule
    \end{tabular}
    }
    \caption{The impact of Outlier Exposure (OE) and contrastive learning $L_{con}$ on NF-based anomaly segmentation. `\# cls' indicates the number of classes for segmentation.}
    \label{table:ablations:contrastive}
\vspace{-3mm}
\end{table}
}

\Cref{table:ablations:contrastive} reveals the critical role of both Outlier Exposure (OE) and contrastive learning in boosting the performance of NF-based anomaly segmentation. Unsupervised variants, trained only on inliers (rows 1-2), struggle to establish a discriminative boundary, yielding poor results. In contrast, introducing outliers during training via any OE strategy (rows 3–7) provides a significant performance improvement across all metrics. This demonstrates that exposing the model to OoD samples is essential to enhance NF's discriminative power and to learn a better representation of the normal data distribution.
However, the method of incorporating outliers is paramount. While discriminative training using standard segmentation losses (rows 3–4) or minimizing outlier log-likelihood (row 5) improves results, these approaches are surpassed by the integration of a contrastive loss ($L_{con}$) in the latent space (rows 6–7). Our full model which combines segmentation and contrastive learning (row 7) achieves the best performance.

The superiority of the contrastive learning approach stems from its ability to directly address a core weakness in standard maximum likelihood training. By functioning as a powerful regularizer, the contrastive loss penalizes the model for mapping outlier features to high-likelihood regions of the latent space typically occupied by inliers. This enforces a more separable latent space, ensuring that the likelihood scores for normal and anomalous samples are more distinct and reliable.

{
\setlength{\textfloatsep}{5pt}
\begin{table}[t]
    \centering
    \resizebox{\columnwidth}{!}{%
    \begin{tabular}{@{}c|c|cc@{}}
    \toprule
    Method & $L_{con}$ & AUPRC $\uparrow$ & $\text{FPR}_{95}$ $\downarrow$ \\
    \midrule
    FastFlow \cite{yu_fastflow_2021} & \xmark & $29.1\pm1.2$ & $41.7 \pm 1.4$   \\
    UFlow \cite{tailanian_u-flow_2023} & \xmark & $20.7 \pm 1.6$ & $66.7\pm10.6$    \\
    FastFlow & \cmark & $40.8\pm2.8$ (+11.7) & $84.9\pm10.4$ (+43.2)    \\
    UFlow & \cmark & $48.7\pm2.3$ (+28.0) & $49.3\pm10.9$ (-17.4)   \\
    \bottomrule
    \end{tabular}
    }
    \caption{The impact of contrastive learning on existing unsupervised NF-based methods.}
    \label{table:ablations:contrastive_mvtec}
\end{table}
}

To demonstrate the broader applicability of our hybrid training objective, we integrated our OE-based contrastive learning framework into two existing unsupervised normalizing flow (NF) methods: FastFlow \cite{yu_fastflow_2021} and UFlow \cite{tailanian_u-flow_2023}. The results in \Cref{table:ablations:contrastive_mvtec} confirm that our approach serves as a powerful enhancement, significantly improving the core anomaly detection capabilities of both models.
Specifically, integrating our method boosts the AUPRC score by 11.7 points for FastFlow and a remarkable 28.0 points for UFlow. This highlights a substantial gain in discriminative power. While this enhancement introduces a trade-off for FastFlow by increasing its false positive rate, UFlow's performance improves drastically, with its FPR$_{95}$ score dropping by 17.4 points. These findings validate that our framework is not model-specific and can be effectively used to augment existing NF-based methods, enhancing their ability to distinguish between normal and anomalous data.

\subsubsection{Impact of various pre-trained encoders}
{
\setlength{\textfloatsep}{5pt}
\begin{table}[t]
    \centering
    \resizebox{\columnwidth}{!}{%
    \begin{tabular}{@{}c|cc|cc@{}}
    \toprule
    \multirow{2}{*}{Encoder} & \multicolumn{2}{c|}{FS-L\&F} & \multicolumn{2}{c}{Road Anomaly} \\
    \cline{2-5} & AUPRC $\uparrow$ & $\text{FPR}_{95}$ $\downarrow$ & AUPRC $\uparrow$ & $\text{FPR}_{95}$ $\downarrow$ \\
    \midrule
    EVA-02-L-Rein (Cityscapes) \cite{fang2024eva,wei_stronger_2024} &
    55.5 & 43.9 & 83.3 & 23.0 \\
    Swin-L (ImageNet-1k) \cite{liu_swin_2021} & 
    84.5 & 12.2 & 83.6 & 10.8 \\
    Swin-L (Cityscapes) \cite{liu_swin_2021} & 
    69.5 & 5.6 & 87.0 & 9.6 \\
    DINO-ViT-B (ImageNet-1k) \cite{caron_emerging_2021} & 
    15.0 & 39.2 & 72.8 & 22.7 \\
    DINOv2-B (LVD-142M) \cite{oquab_dinov2_2023} & 
    81.0 & 1.9 & 94.2 & 5.1 \\
    DINOv2-B-Rein (Cityscapes) \cite{oquab_dinov2_2023,wei_stronger_2024} & 
    80.7 & 1.4 & 90.6 & 5.0 \\
    DINOv2-L ((LVD-142M)) \cite{oquab_dinov2_2023} & 
    85.2 & 0.7 & 92.3 & 4.8 \\
    DINOv2-L-Rein (Cityscapes) \cite{oquab_dinov2_2023,wei_stronger_2024} & 
    86.1 & 0.6 & 92.0 & 4.0 \\
    \bottomrule
    \end{tabular}
    }
    \caption{Ablations analyzing the effect of different encoders.}
    \label{table:ablations:backbone}
\vspace{-3mm}
\end{table}
}

Our ablation study on various feature extractors, detailed in \cref{table:ablations:backbone}, shows that the quality and scale of the backbone's pre-training plays a critical role in the performance of FlowCLAS.
The importance of a diverse pre-training dataset is illustrated by comparing DINO-ViT-B and DINOv2-B. Despite sharing nearly identical architectures, the DINOv2-B model, pre-trained on the large-scale LVD-142M dataset, achieves 81.0 AUPRC, whereas its ImageNet-1k-trained counterpart scores a mere 15.0. Likewise, the performance gap between DINOv2-L-Rein and the weaker EVA-02-L-Rein is primarily attributable to their different pre-training data. This confirms that a rich pre-training dataset is essential for building a feature space with the discriminative semantic information required for anomaly discrimination.
In contrast, the impact of task-specific fine-tuning is secondary and can even be detrimental. While applying Rein fine-tuning to DINOv2-L yields a modest performance boost (+0.9 AUPRC), this trend is not universal. The Swin-L model's performance deteriorates significantly after being fine-tuned on Cityscapes (84.5 vs. 69.5 AUPRC). This suggests that while targeted fine-tuning can provide marginal improvements, it introduces a substantial risk of overfitting to the narrow in-distribution data, thereby degrading the generalizable representations vital for normalizing flows.
These findings underscore that a powerful feature extractor is a crucial ingredient for good performance. However, rich features alone are insufficient to guarantee effective anomaly segmentation as demonstrated in \Cref{table:ablations:contrastive}. It is only when these powerful features are guided by our contrastive objective that their full potential is unlocked, significantly boosting performance. This dependence on feature quality is not a limitation but a core strength of our framework’s post-hoc design. FlowCLAS is explicitly designed to leverage the improving capabilities of off-the-shelf vision foundation models, ensuring its relevance and performance will continue to scale with future advancements in representation learning.

\subsubsection{Effect of mask-based anomaly score smoothing}
{
\setlength{\textfloatsep}{5pt}
\begin{table}[t]
    \centering
    \resizebox{\columnwidth}{!}{%
    \begin{tabular}{@{}c|c|cc|cc|cc@{}}
    \toprule
    \multirow{2}{*}{Method} & \multirow{2}{*}{Smooth.} & \multicolumn{2}{c|}{ALLO} & \multicolumn{2}{c|}{FS-L\&F} & \multicolumn{2}{c}{Road Anomaly} \\
    \cline{3-8} & & AUPRC $\uparrow$ & $\text{FPR}_{95}$ $\downarrow$ & AUPRC $\uparrow$ & $\text{FPR}_{95}$ $\downarrow$ & AUPRC $\uparrow$ & $\text{FPR}_{95}$ $\downarrow$ \\
    \midrule
    RbA \cite{nayal_rba_2023} & \xmark & 50.5 & 75.5 & 70.8 & 6.3 & 85.4 & 6.9 \\
    UNO \cite{delic_outlier_2024} & \xmark & 80.8 & 5.4 & 81.8 & 1.3 & 88.8 & 6.4 \\
    FlowCLAS (DINOv2-L-Rein) & \xmark & 86.7 & 11.6 & 86.1 & 0.6 & 92.0 & 4.0 \\
    RbA \cite{nayal_rba_2023} & \cmark & 47.2 & 77.0 & 73.7 & 7.5 & 85.8 & 6.7 \\
    UNO \cite{delic_outlier_2024} & \cmark & 81.1 & 9.1 & 82.6 & 1.3 & 87.8 & 10.6 \\
    FlowCLAS (DINOv2-L-Rein) & \cmark & 88.4 & 6.6 & 88.8 & 0.7 & 93.0 & 3.3 \\
    \bottomrule
    \end{tabular}
    }
    \caption{Analysis of the effect of mask-based anomaly score smoothing on anomaly segmentation performance. `Smooth.' indicates whether the method incorporated score smoothing.}
    \label{table:ablations:smoothing}
\vspace*{-4mm}
\end{table}
}

Our analysis in \Cref{table:ablations:smoothing} investigates the efficacy and generalizability of our mask-based score smoothing module. To study whether this module provides a universal advantage, we also apply it to recent Mask2Former-based methods, RbA and UNO.
The results reveal a dichotomy. For FlowCLAS, the smoothing module provides consistent and significant performance gains across all metrics. On the ALLO dataset, it boosts the AUPRC by +1.7 points while cutting the FPR$_{95}$ nearly in half (from 12.0 to 6.6). In contrast, when this same module is applied to RbA and UNO, the results are inconsistent and often detrimental. The negative impact is most pronounced for UNO on the Road Anomaly dataset, where smoothing degrades performance and causes the FPR$_{95}$ to spike from 6.4 to 10.6.
This discrepancy strongly suggests there is a mismatch between the predicted masks from SAM 2 and Mask2Former, leading to erroneous score averaging and performance degradation when applying our refinement to RbA and UNO.

\section{Conclusion}
\label{sec:conclusion}

In this work, we address a fundamental limitation of normalizing flows (NFs) in anomaly segmentation for complex, dynamic scenes: their struggle to model highly multi-modal distributions and their tendency to assign high likelihood to out-of-distribution samples. We introduce FlowCLAS, a novel framework that bridges the gap between generative density estimation and discriminative learning. By integrating a contrastive learning objective, which leverages outlier exposure, into a classic NF pipeline, FlowCLAS learns a latent space where normal and anomalous features are explicitly separated. The efficacy of this hybrid approach is proven by our extensive experiments, where FlowCLAS establishes a new state-of-the-art on four challenging robotics benchmarks. Furthermore, our ablations confirm that this contrastive objective is superior to other outlier-based training strategies and that the framework serves as a generalizable enhancement for existing NF methods for robotics applications, validating the robustness and broad applicability of our framework.

\clearpage
{
    \small
    \bibliographystyle{ieeenat_fullname}
    \bibliography{references}
}

\clearpage
\setcounter{page}{1}
\maketitlesupplementary

\section{Methodology}
\subsection{Normalizing flow architecture details}
\label{sec:suppl:nf_details}

\begin{figure*}[htb]
  \centering
   \includegraphics[width=0.8\textwidth]{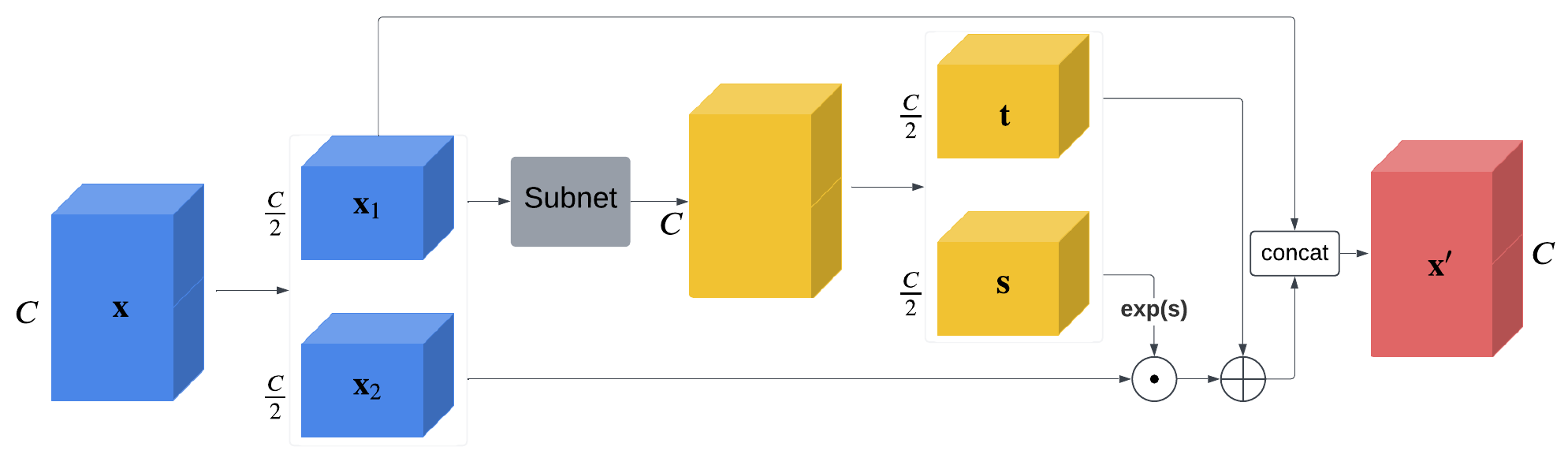}
   \caption{The feature or latent maps $\xb$ are bisected along the channel dimension. The first half $\xb_1$ is processed by a learnable module, $\operatorname{subnet}$ which generates affine transformation parameters $\mathbf{s}$ and $\mathbf{t}$. These parameters are then applied to the second half $\xb_2$, yielding a transformed $\xb_2^{\prime}$. The final output is obtained by concatenating $\xb_1$ and $\xb_2^{\prime}$ along the channel axis.}
\label{fig:suppl:coupling}
\end{figure*}

\begin{figure}[htb]
  \centering
   \includegraphics[width=\linewidth]{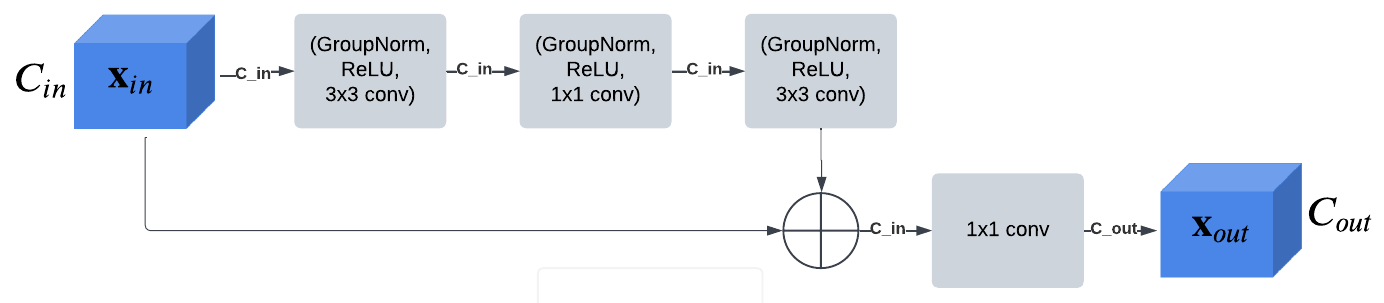}
   \caption{The residual network used in the affine coupling layer, denoted as `subnet' in \cref{fig:suppl:coupling}.}
\label{fig:suppl:subnet}
\end{figure}

This section describes the structure of the affine coupling operation $\operatorname{Affine}$ \cite{dinh_density_2017} used in FlowCLAS which incorporates 2D subnets following \cite{yu_fastflow_2021}. As depicted in \cref{fig:suppl:coupling}, the operation bisects the input $\xb$ along the channel dimension. The first half serves as input to a subnetwork that generates affine transformation parameters which are applied to transform the second half. This process can be formalized as follows:

\begin{equation}
\begin{aligned}
    \xb_1, \xb_2 &= \operatorname{split} \left( \xb \right) \\
    \mathbf{s}, \mathbf{t} &= \operatorname{split} \left( \operatorname{subnet} \left( \xb_1 \right) \right) \\
    \xb_2^{\prime} &= \xb_2 \cdot \exp \left( \mathbf{s} \right) + \mathbf{t} \\
    \xb^{\prime} &= \operatorname{concat} \left(\xb_1, \xb_2^{\prime} \right)
\end{aligned}
\end{equation}
where $\operatorname{subnet}$ is a residual network described in \cref{fig:suppl:subnet}

\subsection{Mask-based Anomaly Score Smoothing}
\label{sec:suppl:score_smoothing}

A detailed pseudo-code of the mask-based anomaly score smoothing pipeline can be found in \cref{alg:smooth_scores}.

\begin{algorithm}[htb]
  \caption{Mask-based Anomaly Score Smoothing}
  \label{alg:smooth_scores}
  \begin{algorithmic}[1]
      \State $masks \gets \texttt{SAM 2}(I)$ \Comment{Segment regions using SAM 2}
      \For{mask $M$ in $masks$}
        \State $(h, bins) \gets \texttt{Histogram}(scores_M, N)$
        \State $bin^* \gets \arg\max_{b}~h(b)$
        \State $score^* \gets \texttt{Mean}(\{s \in bin^*\})$
        \For{pixel $p$ in $M$}
            \If{$\operatorname{score}(p)$ not in $bin^*$}
                \State $S(p) \gets score^*$
            \EndIf
        \EndFor
      \EndFor
      \State \Return $scores$
  \end{algorithmic}
\end{algorithm}

\section{Experiments \& results}

\subsection{Threshold bins for binary classification metrics}
\label{sec:suppl:metric_bins}

Conventional anomaly segmentation test sets, due to their relatively small size, allow each pixel-level anomaly score to function as an individual threshold bin, enabling tractable precise metric calculation. However, to overcome memory limitations and enable online processing with the extensive ALLO test set \cite{leveugle_allo_2024}, we employ predefined threshold bins instead of the traditional method of post-processing all the scores simultaneously. Furthermore, to ensure compatibility with these bins, we constrain the anomaly scores to the $[0,1]$ interval by applying the $\operatorname{sigmoid}(\cdot)$ function prior to metric computation.

\subsection{COCO images for ALLO experiments}

For the ALLO experiments, the backgrounds of COCO images \cite{lin_microsoft_2014} were excluded during training and not used to construct the set $\mathcal{B}$ described in \cref{sec:methodology}. This decision was made to avoid confusion between the predominantly black space background and the colorful real-world backgrounds present in COCO images.

\subsection{Fine-tuning with Rein}

We used Rein \cite{wei_stronger_2024} to fine-tune DINOv2-{B,L} for inlier semantic segmentation on ALLO and Cityscapes. This process follows the semantic segmentation pre-training procedure performed by the supervised baselines. The fine-tuning was conducted with a batch size of 8, a learning rate of $6 \times 10^{-5}$ using polynomial decay, the AdamW optimizer, and 40,000 training iterations.

\subsection{Extra visualizations}

\begin{figure*}[htb]
  \centering
   \includegraphics[width=0.9\textwidth]{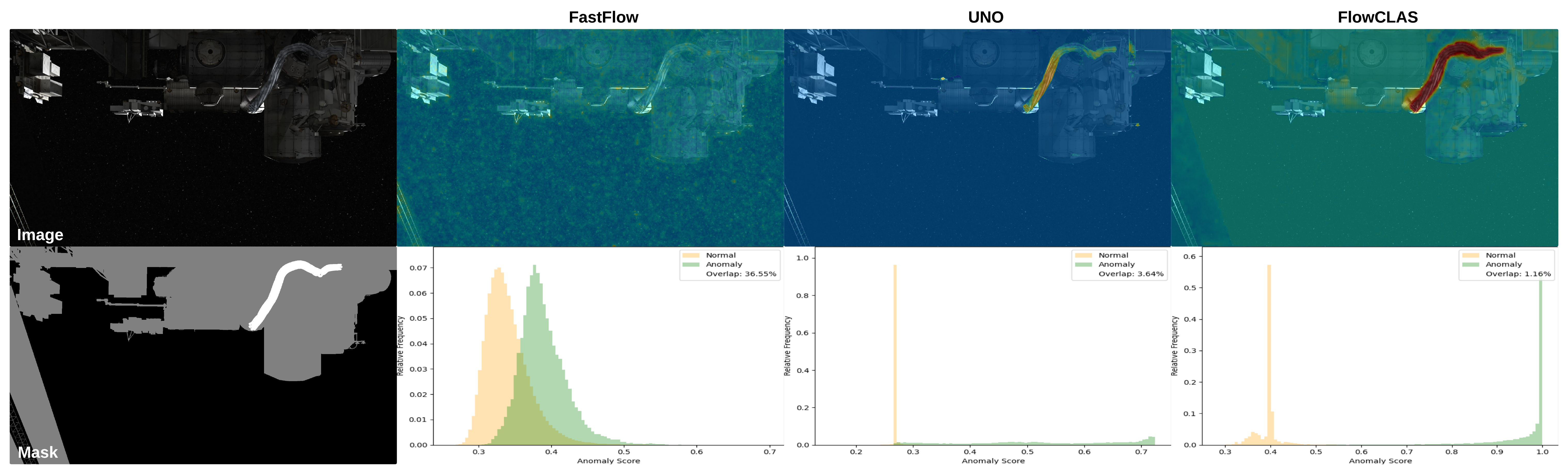}
   \caption{Predicted heatmaps (top) and anomaly score histograms (bottom) for an ALLO test image. While the leading unsupervised method, FastFlow \cite{yu_fastflow_2021}, fails to detect the anomalous cable, and the supervised SOTA, UNO \cite{delic_outlier_2024}, detects a part of the main body, FlowCLAS provides a more complete segmentation that captures the entire object structure. The histograms below visually illustrate this, showing the superior anomaly score separation between the two classes achieved by FlowCLAS compared to both methods.}
\label{fig:suppl:allo_results}
\end{figure*}

\begin{figure*}[htb]
    \centering
    \begin{subfigure}[t]{0.495\textwidth}
        \centering
        \includegraphics[width=\textwidth]{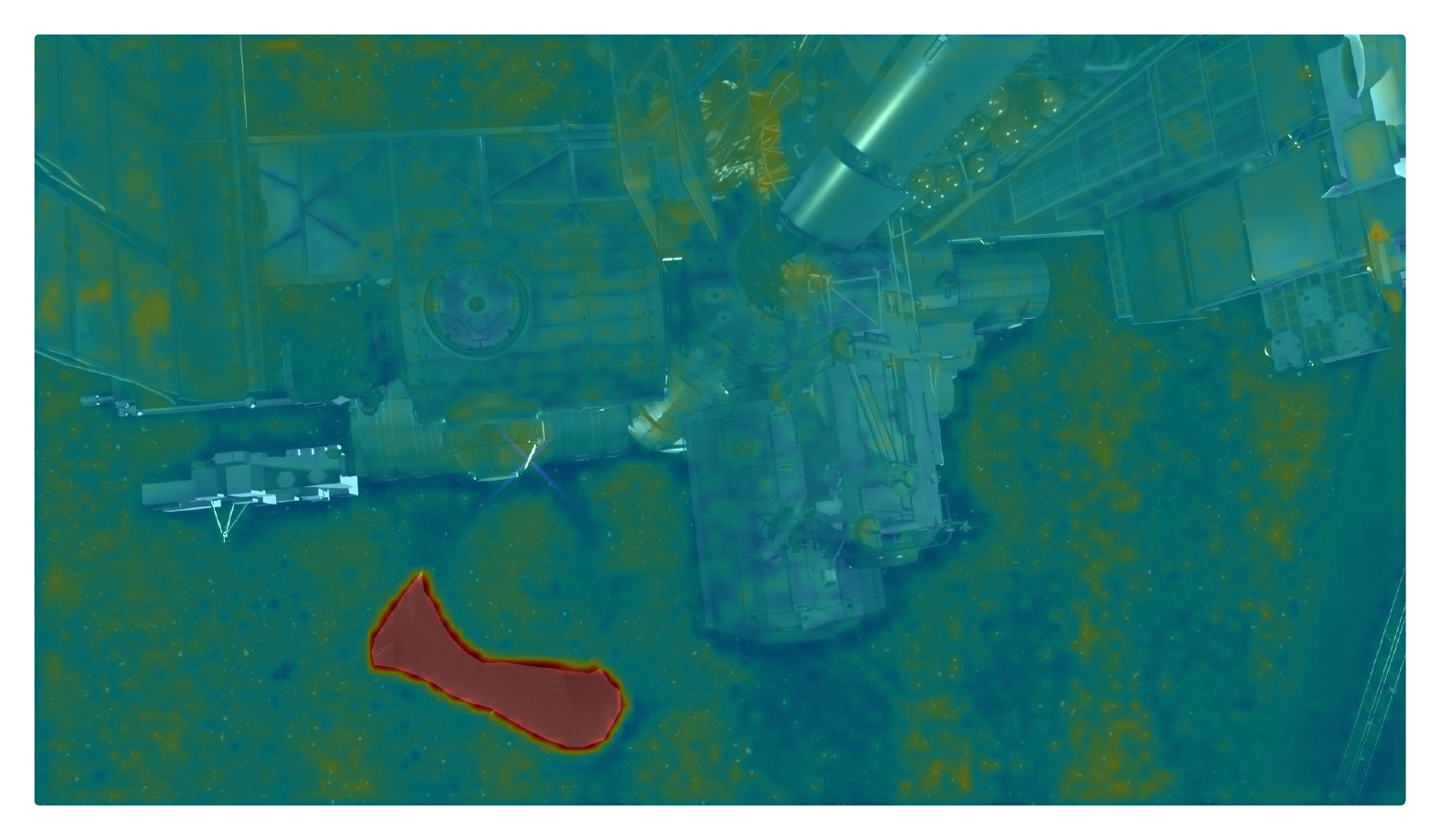}
    \end{subfigure}
    \hfill
    \begin{subfigure}[t]{0.495\textwidth}
        \centering
        \includegraphics[width=\textwidth]{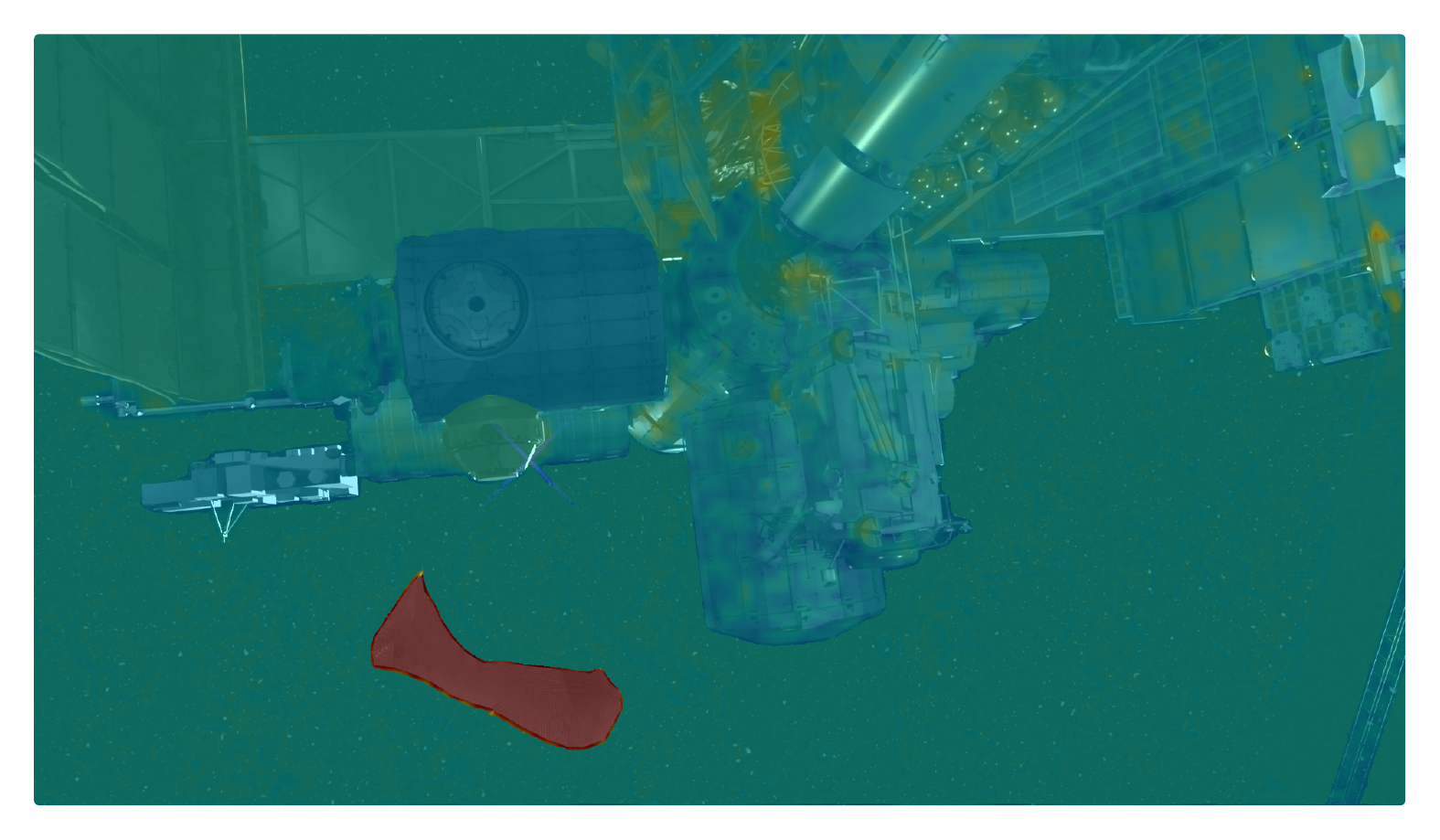}
    \end{subfigure}
\caption{Visual comparison of heatmaps from FlowCLAS without (left) and with (right) mask-based score refinement.}
\label{fig:suppl:sam_comparison}
\end{figure*}

\begin{figure*}[htb]
  \centering
  \subfloat[Fishyscapes Lost\&Found]{%
  \includegraphics[width=0.9\textwidth]{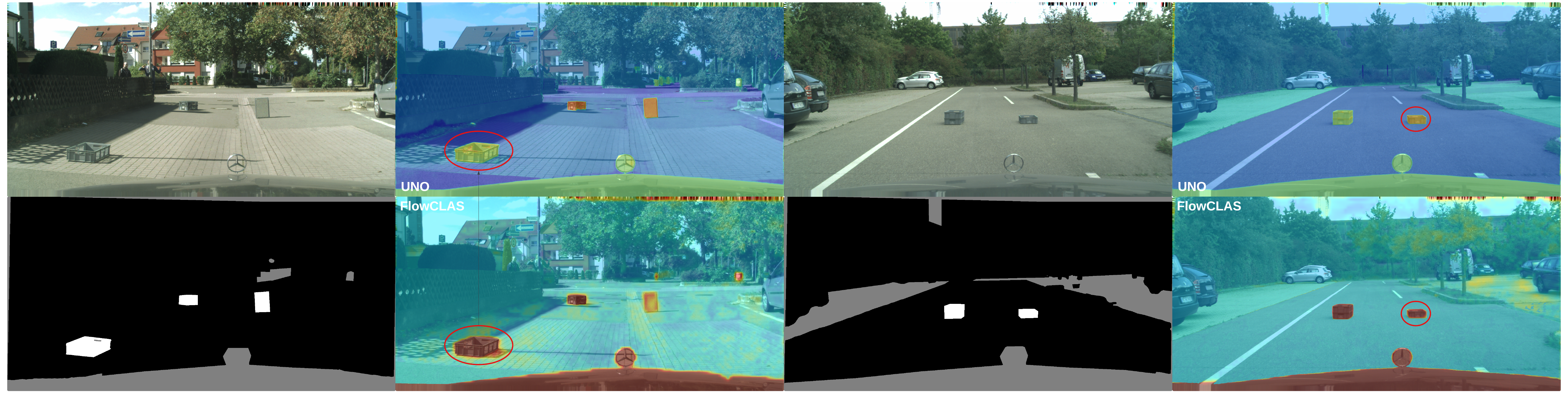}
  \label{fig:suppl:viz:fishy}
}\\
\subfloat[Road Anomaly]{%
       \includegraphics[width=0.9\textwidth]{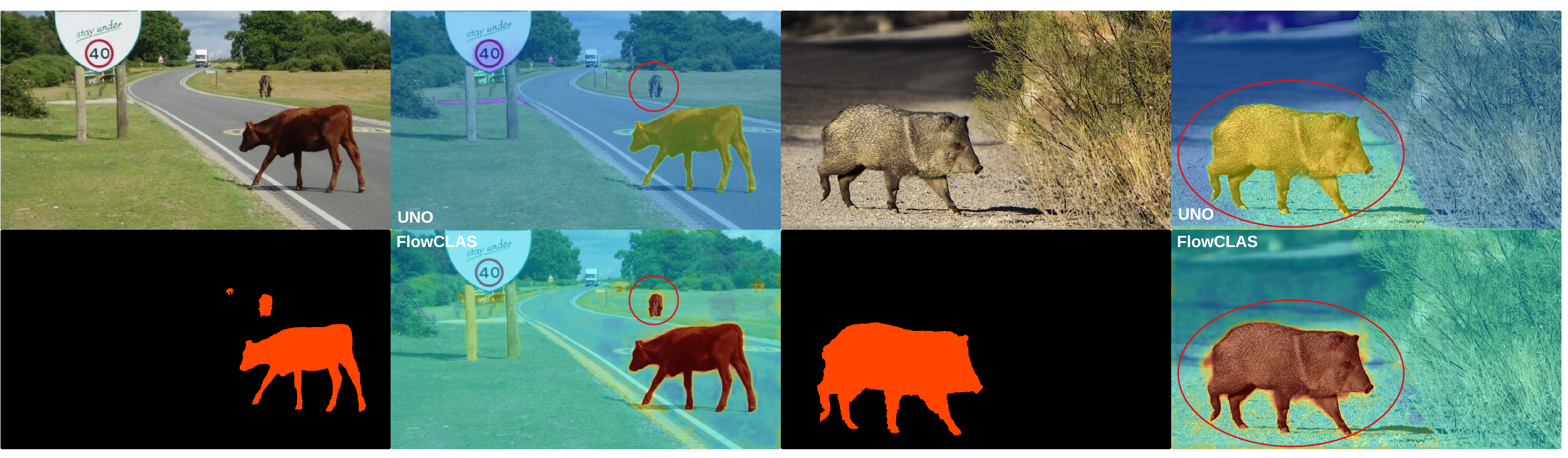}
       \label{fig:suppl:viz:road_anomaly}
}
\caption{Predicted heatmaps from UNO \cite{delic_outlier_2024} and FlowCLAS for examples from the Fishyscapes Lost\&Found \cite{blum_fishyscapes_2019} (top) and Road Anomaly \cite{lis_detecting_2019} (bottom) validation sets.}
\label{fig:suppl:viz}
\end{figure*}

\Cref{fig:suppl:viz} compares anomaly heatmaps predicted by UNO and FlowCLAS on Fishyscapes Lost\&Found (FS L\&F) \cite{blum_fishyscapes_2019} and Road Anomaly \cite{lis_detecting_2019}. In both FS L\&F examples (top), FlowCLAS more accurately detects the baskets than UNO, the previous supervised state-of-the-art. Similarly, FlowCLAS successfully identifies the cow in the Road Anomaly example (bottom-left) and reduces false-positive detections in the road region surrounding the peccary (bottom-right).

\subsection{Additional ablation studies}
\label{sec:suppl:ablations}

\begin{table}[htb]
    \centering
    \resizebox{\columnwidth}{!}{%
    \begin{tabular}{@{}c|cc|cc|cc|c@{}}
    \toprule
    \multirow{2}{*}{$L$} & \multicolumn{2}{c|}{ALLO} & \multicolumn{2}{c|}{FS-L\&F} & \multicolumn{2}{c|}{Road Anomaly} & \multirow{2}{*}{\# of params} \\
    \cline{2-7} & AUPRC $\uparrow$ & $\text{FPR}_{95}$ $\downarrow$
    & AUPRC $\uparrow$ & $\text{FPR}_{95}$ $\downarrow$ & AUPRC $\uparrow$ & $\text{FPR}_{95}$ $\downarrow$ & \\
    \midrule
    8 & 63.0 & 63.1 & 79.2 & 1.7 & 88.5 & 5.4 & 25.0 M \\
    12 & 62.8 & 63.7 & 77.1 & 2.1 & 88.0 & 5.6 & 37.4 M \\
    16 & 63.6 & 64.2 & 80.7 & 1.4 & 90.6 & 5.0 & 49.8 M \\
    \bottomrule
    \end{tabular}
    }
    \caption{Summary of results with varying flow steps $L$ and the total count of trainable parameters. DINOv2-B was used for ALLO and DINOv2-B-Rein was used for FS-L\&F and Road Anomaly.}
    \label{table:suppl:flow_steps}
\end{table}

Our analysis of the flow step count $L$ in FlowCLAS, detailed in \Cref{table:suppl:flow_steps}, reveals a non-monotonic relationship between model depth and performance, highlighting a crucial trade-off between expressivity and optimization.
Theoretically, increasing $L$ enhances the normalizing flow's expressive power. However, our results show that increasing $L$ from 8 to 12 leads to a slight degradation in performance across the road anomaly benchmarks (\eg AUPRC drops from 79.2 to 77.1 on FS-L\&F). This counter-intuitive result suggests that while model capacity increases (from 25.0M to 37.4M parameters), the model may become more difficult to optimize or prone to overfitting. The more complex transformation might model the training inliers too tightly, potentially degrading its ability to generalize for outlier separation without a corresponding increase in expressivity that provides a tangible benefit.
However, increasing the step count further to $L = 16$ consistently yields the best performance across all datasets, achieving the highest AUPRC and lowest FPR$_{95}$ on FS-L\&F and Road Anomaly. This suggests that with a sufficient increase in capacity, the model can finally leverage its enhanced expressivity to learn a superior representation of the complex normal data distribution, overcoming the optimization hurdles observed at $L = 12$. Based on these findings, we selected $L = 16$ for our main experiments to maximize performance, accepting the trade-off in computational cost.

\begin{table}[htb]
    \centering
    \resizebox{\columnwidth}{!}{%
    \begin{tabular}{@{}c|cc|cc@{}}
    \toprule
    \multirow{2}{*}{$\tau$} & \multicolumn{2}{c|}{FS-L\&F} & \multicolumn{2}{c|}{Road Anomaly} \\
    \cline{2-5} & AUPRC $\uparrow$ & $\text{FPR}_{95}$ $\downarrow$ & AUPRC $\uparrow$ & $\text{FPR}_{95}$ $\downarrow$\\
    \midrule
    0.07 & 80.0 & 1.7 & 88.9 & 5.5 \\
    0.10 & 81.1 & 1.4 & 89.4 & 5.2 \\
    0.13 & 80.7 & 1.4 & 90.6 & 5.0 \\
    \bottomrule
    \end{tabular}
    }
    \caption{Summary of results with varying temperature scale parameters $\tau$.}
    \label{table:suppl:temperature}
\end{table}

Our analysis of the temperature scale hyperparameter $\tau$, reveals that FlowCLAS is relatively insensitive to the choice of $\tau$ within the tested range of 0.07 to 0.13. Across both the FS-L\&F and Road Anomaly benchmarks, the AUPRC varies by less than 2 points, indicating that the framework's performance is not overly sensitive to this hyperparameter. Furthermore, the optimal temperature appears to be dataset-dependent. On the FS-L\&F dataset, the performance is non-monotonic, peaking at $\tau = 0.10$, while on the Road Anomaly dataset, both AUPRC and FPR$_{95}$ show a clear monotonic improvement as $\tau$ increases. This suggests that a higher temperature, which ``softens" the contrastive loss, is more beneficial for the Road Anomaly dataset's distribution. Given that $\tau = 0.13$ yields the best overall performance profile---achieving the top AUPRC on Road Anomaly and the best FPR$_{95}$ across both datasets---it was selected as the default value for our main experiments.

\clearpage

\end{document}